\documentclass[a4paper,fleqn]{cas-sc}

\usepackage[numbers,sort&compress]{natbib}

\usepackage[T1]{fontenc}
\usepackage{booktabs, multirow}
\usepackage{amsmath} 
\usepackage{amssymb}
\usepackage{enumitem}
\usepackage{graphicx}
\usepackage{float}
\usepackage{rotating}
\usepackage{xcolor}
\usepackage{colortbl}
\usepackage[ruled,vlined]{algorithm2e}
\usepackage{tikz}
\usetikzlibrary{shapes.geometric, arrows.meta, positioning, fit, calc, shadows}

\definecolor{cTeal}{RGB}{45,135,140}
\definecolor{cBlue}{RGB}{60,105,170}
\definecolor{cAmber}{RGB}{198,140,35}
\definecolor{cGreen}{RGB}{80,150,95}
\definecolor{cGrey}{RGB}{110,115,125}

\makeatletter
\RenewDocumentCommand \printorcid { } {}
\makeatother

\begin{document}
\let\WriteBookmarks\relax
\def\floatpagepagefraction{1}
\def\textpagefraction{.001}

\shorttitle{NTDH for Comprehensive Affective Analysis}
\shortauthors{T. Zhu et al.}

\title[mode=title]{NTDH: Complex Reasoning for Comprehensive Affective Analysis}

\author[1]{Tianlei Zhu}
\author[2]{Zhiwei Liu}[]
\cormark[1]
\author[2]{Yuyan Wang}
\author[1]{Xiao-Yang Liu}
\author[2]{Sophia Ananiadou}

\affiliation[1]{organization={Columbia University},
                city={New York},
                state={NY},
                country={USA}}

\affiliation[2]{organization={Department of Computer Science, The University of Manchester},
                city={Manchester},
                country={United Kingdom}}

\cortext[1]{Corresponding author. E-mail address: \texttt{zhiwei.liu@manchester.ac.uk}}

\begin{abstract}
Comprehensive affective analysis is challenging for two reasons: it spans
heterogeneous prediction tasks with continuous, ordinal, and multi-label
outputs, and affective meaning is context-dependent, requiring conflicting cues
to be reconciled rather than mapped directly to labels. Existing methods learn
this mapping directly and do not model the reconciliation explicitly. We recast
the task as a \emph{complex-reasoning} problem, which yields one output
interface across heterogeneous label spaces and a trajectory over which a
verifiable reward can be optimised; to our knowledge, this is the first such
treatment covering both sentiment and emotion. The obstacle is on the data side:
affective reasoning traces must be synthesised, and generic synthesis is
misaligned with the targets, tolerances, and phenomena of affect, and discards
or leaks its failure cases. We propose \textbf{NTDH}, which addresses these four
failures. \emph{Naturalisation} sets the training answer to the gold label, so
it is correct by construction. A \emph{Tolerance}-aware gate checks each answer
against the task's own scoring margin. \emph{Domain}-aware strategies refine the
reasoning using ideas from affective science. Directional \emph{Hints} report
only the type and direction of an error, without exposing the target. We train
Qwen3-8B with SFT and then GRPO under the same tolerance used for verification
(up to a more permissive construction gate on the multi-label subtask),
and a component ablation quantifies the data-quality effect of each part. Using
$16{,}302$ training records, about $14\times$ fewer than comparable
instruction-tuned systems, the final policy improves over its SFT checkpoint on
five of six official-test metrics and achieves the strongest EI-reg result among
the compared systems, at a Pearson correlation of $0.862$.
\end{abstract}


\maketitle

\section{Introduction}\label{sec:intro}

Affective analysis aims to identify the sentiment, emotions, and affective
intensities expressed in text, providing fundamental signals for applications
such as opinion and stance analysis, emotion-aware retrieval, conversational
agents, mental-health monitoring, and social-media analysis
\cite{zhang2024affectivesurvey}. Despite substantial progress, comprehensive
affective analysis remains challenging for two fundamental reasons. First, it
encompasses heterogeneous prediction tasks. Sentiment and emotion capture
complementary dimensions of affect, while outputs may be represented as
continuous intensities, ordinal categories, or multi-label emotion sets
\cite{munezero2014}. Consequently, heterogeneous labels require unified
representations while preserving their task-specific semantics and evaluation
criteria. Second, affective meaning is inherently context-dependent. Reliable
prediction requires integrating multiple contextual cues, including polarity
shifts (e.g., negation and intensification) \cite{kennedy2006,wilson2009},
figurative language (e.g., irony, sarcasm, and idioms)
\cite{reyes2012,prochnow2024}, and interactions among multiple emotions
\cite{russell1980circumplex,mill2018role,jabreel2019deep}, rather than directly
mapping text to labels. We investigate these challenges across the four
complementary SemEval-2018 subtasks: emotion-intensity regression (EI-reg),
multi-label emotion classification (E-c), valence regression (V-reg), and
ordinal valence classification (V-oc) \cite{mohammad2018semeval}.

Research on affective analysis has evolved from feature-based methods
\cite{turney2002thumbs,pang2002thumbs,hutto2014vader} through pre-trained
language models to instruction-tuned large language models (LLMs), of which
EmoLLM \cite{emollm} is the
closest to our setting; \S\ref{sec:rw-affect} reviews this line in detail.
Nevertheless, existing methods primarily learn a direct mapping from input text
to output labels, without explicitly modelling the reasoning process required to
reconcile contextual and potentially conflicting affective evidence.

Modelling this process explicitly is valuable less for immediate accuracy gains
than for three capabilities that a direct mapping cannot provide. First, it
supplies a single output interface for heterogeneous label spaces: continuous
intensities, ordinal categories, and multi-label emotion sets can all be
expressed as a structured reasoning trace terminating in a natural-language
conclusion, so that affective cue analysis is shared across subtasks and only
the final commitment remains task-specific. This makes a single policy, rather
than one specialised model per output space, feasible for comprehensive
affective analysis. Second, affective labels are verifiable under task-specific
criteria, namely numerical tolerance for regression, exact match for ordinal
classes, and set agreement for multi-label outputs. Verifiability is a
precondition for reward-based optimisation, but a reward requires a space over
which to optimise; an explicit reasoning trace provides that space, whereas a
single scalar prediction offers reinforcement learning nothing to reshape.
Third, an explicit trace renders the judgement inspectable, which is important
both for a task whose annotations are themselves contested and for deployment
settings such as mental-health monitoring, where an unexplained score is not
actionable.

Recent reasoning frameworks combining supervised reasoning traces with
reinforcement learning have demonstrated remarkable success in domains such as
mathematics \cite{deepseekai2025r1,cobbe2021training} and clinical question
answering \cite{huatuogpt-o1}, where rationales can be synthesised by a teacher
model and filtered by a judge \cite{zelikman2022star,hsieh2023distilling}.
However, directly transferring these frameworks to affective analysis remains
challenging. Generated reasoning traces may terminate in incorrect or
inconsistently formatted conclusions, generic verification mechanisms cannot
faithfully evaluate heterogeneous regression and classification tasks,
domain-independent refinement offers limited support for affective phenomena
such as valence shifting, irony, and emotion interactions, and unsuccessful
trajectories are commonly discarded or repaired by revealing the gold answer.
These failures define four requirements for
constructing reliable affective reasoning data: (1)~task-consistent target
representations, (2)~evaluation-aware verification, (3)~domain-informed
reasoning, and (4)~informative guidance for resolving difficult cases without
compromising supervision. Existing reasoning frameworks fail to satisfy these
four requirements simultaneously.

To address these limitations, we propose \textbf{NTDH}, a \emph{quality-aware
reasoning-data synthesis framework} for cross-task affective analysis. Rather
than treating initially generated reasoning traces as ready-to-use supervision,
NTDH controls four dimensions of reasoning-data quality through four
complementary components, one per requirement. \textbf{Naturalisation (N)}
converts heterogeneous task labels into scale-aware natural-language targets
while preserving their task-specific semantics. \textbf{Tolerance-aware
Verification (T)} evaluates generated conclusions using task-specific acceptance
criteria that account for numerical tolerances, categorical decisions, and
multi-label agreement. \textbf{Domain-aware Refinement (D)} revises unsuccessful
reasoning trajectories using affect-specific knowledge, including contextual
valence shifts, figurative language, and interactions among emotions
(\S\ref{sec:methodology}). \textbf{Directional Hints (H)} provide target-derived
corrective feedback without directly disclosing the exact reference answer,
enabling difficult trajectories to be refined while reducing direct answer
leakage.

Using the resulting quality-controlled corpus, we first perform supervised
fine-tuning (SFT) to initialise Qwen3-8B \cite{qwen3} with cross-task affective
reasoning patterns. We subsequently apply Group Relative Policy Optimization
(GRPO) \cite{shao2024deepseekmath,huatuogpt-o1} with task-specific verifiable
rewards, defined by the same tolerance the verifier applies, to further optimise
its reasoning and prediction policy. Together, these components provide a
unified approach to constructing and exploiting reasoning supervision across
heterogeneous affective tasks.

Our contributions are:
\begin{enumerate}[nosep]
  \item \textbf{A unified reasoning formulation for affective analysis.} We
  formulate sentiment, emotion, intensity regression, ordinal prediction and
  multi-label classification as structured generation with an explicit
  reasoning path, rather than as independent text-to-label mappings, obtaining
  one output interface and one verifiable reward across four heterogeneous
  subtasks.
  \item \textbf{NTDH, a quality-aware reasoning pipeline.} Four data-side fixes,
  each grounded in the affect literature: \emph{Naturalisation},
  \emph{Tolerance}-aware verification, \emph{Domain}-aware refinement, and
  directional \emph{Hints}. Only $18.4\%$ of the initial model-generated
  conclusions satisfy the gold tolerance; Naturalisation instead supplies a
  gold-consistent \texttt{<answer>} for every retained SFT trace. Of the half-A
  pool, 5{,}388 verified traces form the SFT set, while the remaining 2{,}762 hard cases
  are routed to RL with half-B. This routing uses all $16{,}302$ instances
  ($5{,}388$ SFT $+$ $10{,}914$ RL). A component ablation
  (\S\ref{sec:ablation}) checks each part: N drives target correctness, T
  removes $48$--$63\%$ of the out-of-tolerance accepts, H eliminates label
  leakage from the fallback, and D helps the fine-grained E-c subtask.
  \item \textbf{Evaluation across four affective tasks.} Using $16{,}302$ total
  SFT and RL records---about $14\times$ fewer than EmoLLM's $234$K instruction
  records---RL-final outperforms the SFT initialisation checkpoint on five of six
  metrics on the official $9{,}201$-instance test set and achieves the strongest
  EI-reg result among the compared systems, while remaining competitive across
  all four subtasks. Component ablations and qualitative error analysis identify
  where performance changes and where rare-emotion recall remains limiting.
\end{enumerate}

\section{Related Work}\label{sec:related}

\subsection{Affective Analysis of Text}\label{sec:rw-affect}

Recent affective-analysis research has shifted from task-specific encoders to
generative language models \cite{zhang2024affectivesurvey}. Transformer encoders
such as BERT \cite{devlin2019bert} and RoBERTa \cite{liu2019roberta}, together
with affect- and social-media-specialised variants including BERTweet
\cite{nguyen2020bertweet} and SentiBERT \cite{yin2020sentibert}, remain strong
supervised baselines, but are typically fine-tuned with a separate prediction
head for each label space. This task-specific paradigm also characterises the
original SemEval-2018 Task~1 systems \cite{mohammad2018semeval}: SeerNet combines
multiple lexical regressors \cite{duppada2018seernet}, whereas NTUA-SLP uses
deep attentive recurrent networks \cite{baziotis2018ntua}.

Work since 2023 has increasingly examined whether LLMs can replace these
separate pipelines with prompting or instruction tuning. A broad evaluation
over 13 sentiment-analysis tasks and 26 datasets finds that LLMs are effective
on simpler settings but still lag specialised models on several complex or
structured tasks \cite{zhang2023realitycheck}. Recent systems therefore target
particular sources of difficulty: THOR elicits reasoning for implicit sentiment
\cite{fei2023reasoning}; InstructERC combines instruction tuning and retrieval for
conversational emotion \cite{lei2023instructerc}; Feng et al.\ compare in-context
learning and task-specific fine-tuning for conversational affect
\cite{feng2024affect}; DECC decomposes emotion-cause reasoning
\cite{wu2024decc}; and TEII develops iterative LLM workflows for cross-lingual
emotion detection \cite{cheng2024teii}.

Other recent work broadens either task coverage or interpretability. EmoLLM
unifies multiple affective tasks through large-scale instruction tuning
\cite{emollm}; EmotionQueen evaluates event, implicit-emotion, intention and
empathetic-response capabilities \cite{chen2024emotionqueen}; and MASIVE moves
beyond a small closed inventory toward open-ended affective-state generation
\cite{deas2024masive}. EmoRationale adds retrieved evidence to multilingual
emotion and intensity predictions \cite{saeedi2025emorationale}, while
cognitive-science-based analyses argue for grounding model design in theories of
emotion and communication \cite{bonard2024cognitive}. Nevertheless, these
approaches mainly address a single affective setting, evaluate general-purpose
LLMs, or generate answers without constructing verifier-controlled reasoning
trajectories across sentiment, emotion and intensity. NTDH addresses this gap
while retaining EmoLLM's unified, instruction-following formulation.

\subsection{Complex Reasoning with LLMs}\label{sec:rw-reasoning}

A second line of work makes LLMs reason step by step. Chain-of-thought (CoT)
prompting elicits intermediate reasoning and improves multi-step tasks
\cite{wei2022cot}, and works even zero-shot \cite{kojima2022zeroshot}. Later
methods sharpen it: self-consistency votes over many sampled chains
\cite{wang2023selfconsistency}. Inference-time reasoning then grew into a
paradigm of its own with OpenAI o1 \cite{jaech2024openai} and DeepSeek-R1
\cite{deepseekai2025r1}.

Reinforcement learning turns reasoning from a prompt into a trained skill.
Reinforcement learning from human feedback (RLHF) can align language models
with human preferences by optimising a learned reward
model; InstructGPT performs this optimisation using PPO
\cite{ouyang2022instructgpt,schulman2017proximal}.
When answers are checkable, a verifier can supply the reward. Verifiers were
first trained for mathematical word problems \cite{cobbe2021training}, and DeepSeek-R1
\cite{deepseekai2025r1} shows that RL with verifiable rewards alone can induce
strong reasoning. Group Relative Policy Optimization (GRPO)
\cite{shao2024deepseekmath} makes this practical by estimating advantages from
groups of samples and dropping the value network; open implementations
\cite{openr1,trl} reproduce the SFT-then-GRPO pipeline.

These recipes need reasoning data, which is usually synthesised by a teacher
model. STaR bootstraps rationales by keeping the ones that reach the right answer
\cite{zelikman2022star}, and step-by-step distillation transfers a teacher's
rationales to a smaller model \cite{hsieh2023distilling}. HuatuoGPT-o1
\cite{huatuogpt-o1} is the closest method to ours: a verifier first guides the
model to synthesise correct CoT traces for SFT, then a second RL stage refines them. The
recurring problem is the \emph{quality} of the synthesised traces. Many reach the
right answer by unfaithful reasoning, or fail; the usual fix is to keep verified
traces and discard the rest, which throws away much of the pool. NTDH targets
exactly this data-quality problem.

\subsection{Positioning: Reasoning Meets Affect}\label{sec:rw-positioning}

These two lines have rarely met. On the affect side, reasoning has so far been
used mostly at inference time, through prompting. THOR chains CoT steps to infer
implicit sentiment \cite{fei2023reasoning}; recent work also adds decomposed reasoning for
emotion-cause extraction \cite{wu2024decc}. None of these synthesises
quality-controlled, RL-ready affective reasoning data, and none trains with a
verifiable reward. On the reasoning side, the verifier-driven recipes above were
built for mathematics and medicine. Their generic ``rethink'' steps are blind to the
cues that govern affect: valence shifters, irony, idioms, the emotion/sentiment
distinction, and the circumplex.

We sit at the intersection. To our knowledge, we are the first to bring the
verifiable-reward reasoning recipe to comprehensive affective analysis through
affect-specific synthesis, verification and refinement. Unlike HuatuoGPT-o1
\cite{huatuogpt-o1}, which discards unconverged cases, and EmoLLM
\cite{emollm}, which is trained with SFT alone, we naturalise the gold label and route
the hard cases into GRPO \cite{shao2024deepseekmath}, so no training instance
is wasted. \S\ref{sec:intro} quantifies this claim.

\section{Reasoning-Path Synthesis}\label{sec:methodology}

We present \textbf{NTDH}, a quality-aware \emph{reasoning-path synthesis}
pipeline that converts raw SemEval-2018 labels into verified,
reasoning-augmented training data. Inspired by HuatuoGPT-o1
\cite{huatuogpt-o1}, we adapt its SFT$\rightarrow$GRPO framework to
comprehensive affective analysis: SFT acquires the structured affective
reasoning and prediction format from verified trajectories, and GRPO further
improves the policy with task-aware rewards. We describe the training stage in
\S\ref{sec:training} and defer all hyperparameters to \S\ref{sec:setup}.

\begin{figure}
\centering
\includegraphics[width=\linewidth]{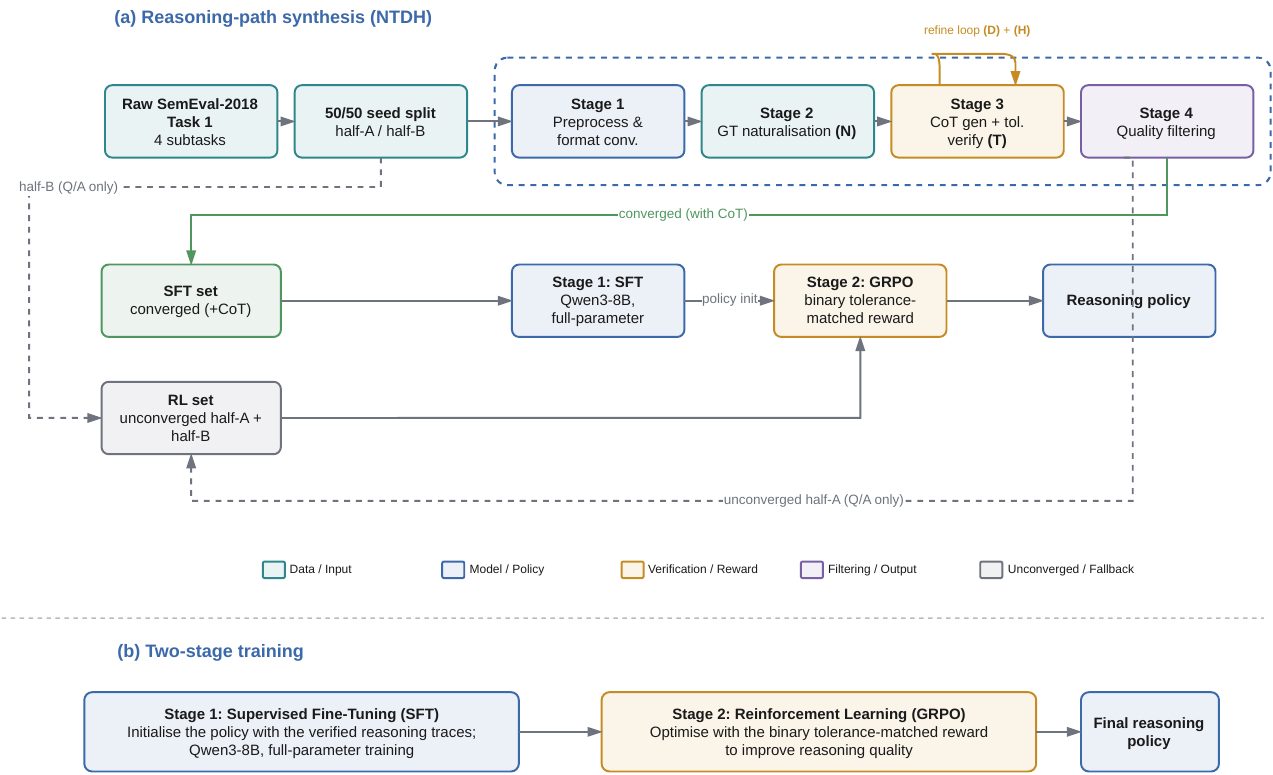}
\caption{Overview of the proposed approach. NTDH turns raw SemEval-2018 data
into verified, reasoning-augmented supervision through four synthesis stages.
Naturalisation~\textbf{(N)}, tolerance-aware verification~\textbf{(T)},
domain-aware refinement~\textbf{(D)} and directional hints~\textbf{(H)} are
realised in Stages~2--3. Converged traces supervise SFT; unconverged hard cases
and half-B supply GRPO, yielding the final affective reasoning policy in one
integrated pipeline.}
\label{fig:pipeline}
\end{figure}

NTDH builds on the EmoLLM Chain-of-thought (CoT) synthesis
pipeline~\cite{emollm} and improves four of its stages, one per letter of the
name:
\begin{itemize}[nosep]
  \item \textbf{(N)~Naturalisation} rewrites each raw label as a scale-aware
        sentence, so the generator, the verifier, and the final supervised
        target all share one natural-language form
        (\S\ref{sec:pipeline}, Stage~2);
  \item \textbf{(T)~Tolerance-aware verification} replaces the original binary
        LLM judge with a deterministic gate matched to each subtask's scoring
        tolerance (\S\ref{sec:task-aware-verify});
  \item \textbf{(D)~Domain-aware refinement} grounds the four rethink strategies
        in the affective-science literature, so the refinement mirrors how affect is
        expressed and resolved (\S\ref{sec:rw-grounding});
  \item \textbf{(H)~Directional Hints} steer a stalled trajectory toward the
        gold answer without ever revealing it (\S\ref{sec:hint-refinement}).
\end{itemize}
Each surviving sample is tagged with a quality tier that drives the SFT/RL
routing (\S\ref{sec:quality-tiers}). The end-to-end flow runs in four stages,
shown in Figure~\ref{fig:pipeline}. We first follow these stages in order
(\S\ref{sec:pipeline}), then detail the task formulation, tolerance-aware gate
(\S\ref{sec:task-aware-verify}), domain-aware refinement and directional hints
(\S\ref{sec:hint-refinement}), and quality-based routing
(\S\ref{sec:quality-tiers}).

\subsection{Data Processing Pipeline}\label{sec:pipeline}

We now walk through the four
pipeline stages of Figure~\ref{fig:pipeline}, from raw SemEval-2018
data to quality-filtered, reasoning-augmented training data; the improvements
above are realised within Stages~2 and~3.

\paragraph{Stage 1: Preprocessing and format conversion.}
We convert the raw SemEval-2018 Task~1 data into a task-specific instruction
that embeds the analysis objective and the expected output format, paired with
the ground-truth label. The encodings differ by subtask: EI-reg names the
target emotion and requests a score in $[0,1]$; V-reg is rescaled to the
symmetric scale $[-1,1]$, so a stored value such as $0.532$ already denotes a
mildly positive valence; V-oc enumerates the seven ordinal classes
($-3$\,=\,very negative, \ldots, $3$\,=\,very positive); and E-c uses a fixed
index map over the eleven emotions (1.~joy, \ldots, 11.~trust; 0.~neutral when
none applies; the ordering is fixed in the released code and is not the
alphabetical listing of \S\ref{sec:task-formulation}), applied identically in
the gold target, the prompt, and the exact-set reward.

\paragraph{Stage 2: Ground-truth naturalisation.}
We convert each raw label into a fluent, scale-aware sentence with a
per-subtask template (Table~\ref{tab:gt-naturalisation}). The supervised
\texttt{<answer>} target is set to this naturalised \emph{ground truth}, not to
the model's (possibly drifted) rewritten response; this is the single fix behind
the answer-source correctness reported in \S\ref{sec:ablation}. Because the
conversion runs before CoT generation, the verifier compares model outputs
against the same natural-language form as the final targets, removing one source
of format mismatch.

\begin{table}[t]
\centering\small
\caption{Ground-truth naturalisation templates. Each subtask's raw
label is converted into a sentence that explicitly states the task
scale.}
\label{tab:gt-naturalisation}
\begin{tabular}{@{}lp{3.0cm}p{8.0cm}@{}}
\toprule
\textbf{Task} & \textbf{Raw} & \textbf{Naturalised sentence} \\
\midrule
EI-reg & \texttt{0.562}
  & \emph{Based on the text, the anger emotion has an intensity of 0.562
    on a scale from 0 (lowest) to 1 (highest).} \\[3pt]
V-reg & \texttt{0.532}
  & \emph{The sentiment intensity of this text is 0.532 on a scale from
    $-$1 (most negative) to 1 (most positive).} \\[3pt]
V-oc & \texttt{2}
  & \emph{Intensity Class: 2: moderately positive mental state can be inferred.} \\[3pt]
E-c & \texttt{7.\,anticipation,} \newline \texttt{9.\,optimism}
  & \emph{The text's emotional tone is: 7.~anticipation, 9.~optimism.} \\
\bottomrule
\end{tabular}
\end{table}

\begin{figure}

\centering
\includegraphics[width=0.72\linewidth]{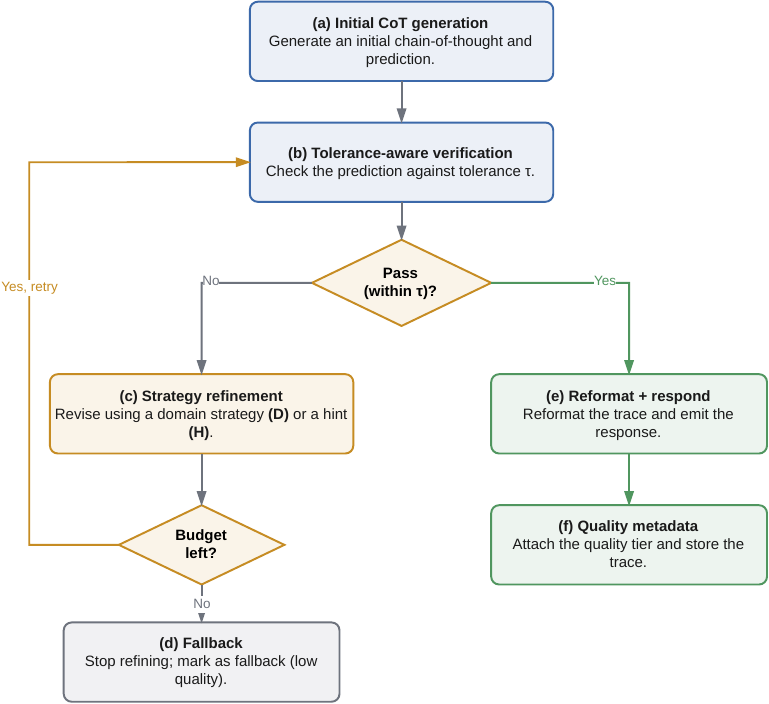}
\caption{Inner loop of Stage~3 for a single sample. The refinement
cycle (c\,$\to$\,b) may iterate up to $D \times A$ times (where $D$ is
the search depth and $A$ the number of restart attempts) before the
fallback path~(d) is taken.}
\label{fig:cot-inner}
\end{figure}

\paragraph{Stage 3: CoT generation with verification.}
This is the core stage, summarised in Figure~\ref{fig:cot-inner}. A generator
LLM produces an initial chain of \emph{Inner Thinking} steps and a \emph{Final
Conclusion}, which the deterministic tolerance gate of
\S\ref{sec:task-aware-verify} checks against the naturalised ground truth. On
failure, the pipeline enters a two-level search: an outer loop of $A$~fresh
restarts (default $A{=}3$) and an inner loop of $D$~refinement rounds (default
$D{=}3$), each round applying one of the four domain-aware strategies of
\S\ref{sec:hint-refinement} and re-verifying. When the budget is exhausted, the
fallback of \S\ref{sec:hint-refinement} applies: hint refinement, allow failure,
or label leakage. On convergence, the chain is reformatted into fluent,
step-by-step reasoning, and a quality tier is assigned from the verification
trajectory (\S\ref{sec:quality-tiers}).

\paragraph{Stage 4: Quality filtering and routing.}
The final stage routes samples as in \S\ref{sec:quality-tiers}: converged half-A
samples (with CoT) form the SFT set; half-B and the unconverged half-A samples
(question/answer only) form the RL set.

\subsection{Task Formulation}\label{sec:task-formulation}

\begin{figure}
\centering
\includegraphics[width=\linewidth]{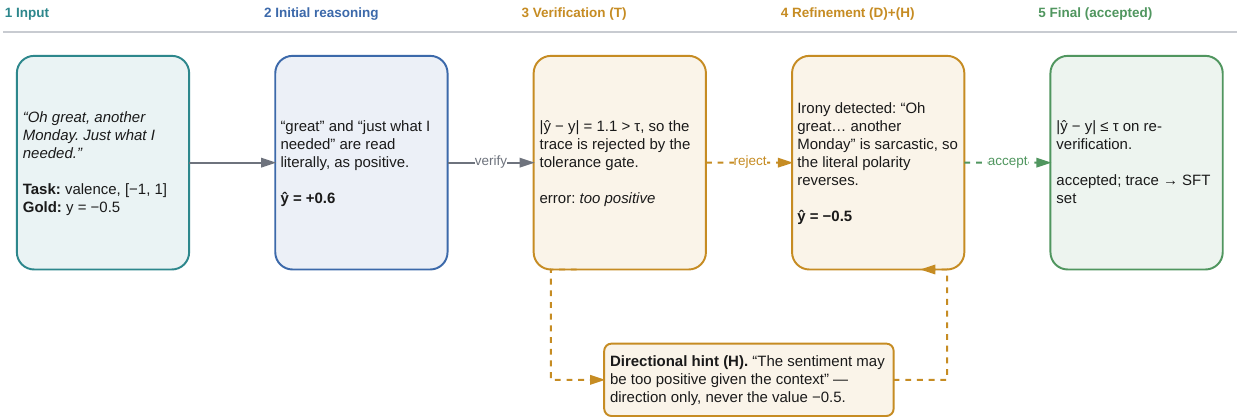}
\caption{A worked example of one NTDH synthesis loop (illustrative). An ironic
tweet is first read literally as positive. The tolerance-aware verifier
\textbf{(T)} computes the task-specific error and rejects the trace; the
directional hint \textbf{(H)} reports only that the prediction is too
positive, without exposing the gold value; and \emph{Backtracking} \textbf{(D)}
then revisits the literal reading and detects the irony. The corrected trace
passes verification and enters the SFT set.}
\label{fig:worked-example}
\end{figure}

We treat all four subtasks as a single conditional generation problem. Let $x$
denote an input tweet paired with a task-specific instruction, and let the
policy $\pi_\theta$ map $x$ to a structured output containing an explicit
reasoning trace and a final answer $\hat{y}$. The subtasks differ only in the
form of $\hat{y}$ and the criterion by which it is scored:
\begin{itemize}[nosep]
  \item \textbf{EI-reg} (emotion-intensity regression): given a tweet and a
        target emotion $e\in\{$anger, fear, joy, sadness$\}$, predict an
        intensity $\hat{y}\in[0,1]$.
  \item \textbf{V-reg} (valence regression): predict a real-valued sentiment
        $\hat{y}\in[-1,1]$.
  \item \textbf{V-oc} (ordinal valence classification): predict one of seven
        ordinal classes $\hat{y}\in\{-3,\dots,3\}$.
  \item \textbf{E-c} (multi-label emotion classification): predict the subset
        $\hat{y}\subseteq\mathcal{E}$ of present emotions, with
        $\mathcal{E}=\{$anger, anticipation, disgust, fear, joy, love, optimism,
        pessimism, sadness, surprise, trust$\}$ (or neutral when none applies).
\end{itemize}
A prediction is counted correct under a task-matched strict criterion $\tau$: absolute
error $|\hat{y}-y|\le0.05$ for the regression subtasks, exact class match for
V-oc, and exact set match for E-c (Table~\ref{tab:verify-thresholds}). This
$\tau$ defines the GRPO accuracy reward (\S\ref{sec:training}) and the strict
accuracy reported in our evaluation. The construction-time gate reuses it for
the regression and ordinal subtasks, but is more permissive for E-c, where it
accepts a trace at $F_1 \geq 0.7$ against the gold set
(\S\ref{sec:task-aware-verify}). Final reporting additionally follows the
official SemEval metrics: Pearson correlation for EI-reg, V-reg, and V-oc, and
Jaccard, micro-F1, and macro-F1 for E-c (\S\ref{sec:setup}). In
Algorithm~\ref{alg:ntdh}, component
\textbf{(D)} samples one of four domain-aware refinement operations:
\emph{Backtrack} revisits an earlier interpretation, \emph{Explore} tests an
alternative cue path, \emph{Verify} checks the draft against textual evidence,
and \emph{Correct} repairs a local label or intensity error; their affective
grounding is detailed in \S\ref{sec:rw-grounding}. Here \emph{Verify} is a
reasoning-revision strategy within \textbf{(D)}, distinct from the deterministic
accept/reject gate \textbf{(T)}. The per-sample synthesis
procedure is illustrated on a concrete tweet in
Figure~\ref{fig:worked-example}. The fifth SemEval-2018 subtask, emotion-intensity
ordinal classification (EI-oc), is a coarse re-binning of the EI-reg signal and
is omitted as redundant.

In Algorithm~\ref{alg:ntdh}, $\hat{y}$ is the generator's candidate conclusion
and is used only to verify whether the associated reasoning chain satisfies the
task-specific construction criterion. Once the chain converges, the
naturalised gold answer $y$---rather than $\hat{y}$---is written to the
supervised \texttt{<answer>} field.

\begin{algorithm}[t]
\DontPrintSemicolon
\caption{NTDH reasoning-path synthesis (single sample)}\label{alg:ntdh}
\KwIn{tweet+instruction $x$; gold answer $y$; tolerance $\tau$; max restarts $A$, depth $D$}
\KwOut{reasoning trace and naturalised gold answer (SFT), or \textnormal{unconverged} (RL)}
$y \leftarrow \textsc{Naturalise}(y)$ \tcp*{(N) scale-aware gold sentence}
$(\mathrm{chain},\hat{y}) \leftarrow \textsc{GenerateCoT}(x)$\;
\For{$a \leftarrow 1$ \KwTo $A$}{
  \For{$d \leftarrow 1$ \KwTo $D$}{
    \If{$\hat{y}$ \textnormal{matches} $y$ \textnormal{within} $\tau$ \textnormal{(T)}}{
      \Return $\textsc{Reformat}(\mathrm{chain}),\ y$ \tcp*{gold $y$ supervises the answer}
    }
    \tcp{(D) sample one domain-aware refinement operation}
    $s \leftarrow \textsc{SampleDomainStrategy}(\{\textsc{Backtrack},\textsc{Explore},\textsc{Verify},\textsc{Correct}\})$\;
    $h \leftarrow \textsc{DirectionalHint}(\hat{y},y)$ \tcp*{(H) label-free guidance}
    $(\mathrm{chain},\hat{y}) \leftarrow \textsc{Refine}(x,\mathrm{chain},s,h)$\;
  }
  reset $\mathrm{chain}$\;
}
\Return \textnormal{unconverged} \tcp*{no CoT $\rightarrow$ RL set}
\end{algorithm}

\subsection{Tolerance-Aware Verification}\label{sec:task-aware-verify}

In the original EmoLLM construction pipeline~\cite{emollm}, after the generator
produces a CoT and final conclusion, a separate judge LLM receives that
conclusion together with the reference answer and is prompted to return
\texttt{True} (semantically consistent) or \texttt{False} (inconsistent). This
judgement is a data-construction filter---it decides whether the trace is
retained or refined---rather than the model's task prediction. Because the
binary prompt does not define task scales or numerical tolerances, it can reject
a valid regression near-match (e.g.\ 0.55 versus 0.56) or accept a value outside
the evaluation tolerance. In the LLM-judge configuration we replaced, such
false acceptance occurred for
$48$--$63\%$ of the out-of-tolerance regression responses presented to the
judge, allowing incorrect targets into SFT.

NTDH separates format interpretation from the final decision. The judge LLM is
retained only to parse heterogeneous natural-language conclusions and for
logging; a deterministic tolerance-aware verifier \textbf{(T)} compares the
parsed prediction with the gold answer under the task-specific construction
criterion in Table~\ref{tab:verify-thresholds}. If the trace fails, the verifier
passes only the signed error category---not the gold value---to the directional
hint module \textbf{(H)}. Thus, in Figure~\ref{fig:worked-example}, \textbf{(T)}
computes $|0.6-(-0.5)|=1.1>0.05$ and rejects the trace, whereas \textbf{(H)}
communicates only ``too positive''. The domain-aware strategy \textbf{(D)}
then determines how to revise the reasoning, here by backtracking to inspect
irony. For E-c only, the construction gate accepts $F_1\geq0.7$ so a mostly
correct reasoning trace can supervise SFT, whereas the GRPO accuracy reward
requires an exact label-set match. Because Naturalisation pairs an accepted
trace with the complete gold label set, a trace that passes this permissive
gate may not explicitly justify every label in its supervised answer. This
reasoning--answer mismatch is a limitation of the present construction and is
discussed in \S\ref{sec:discussion}.

\begin{table}[t]
\centering\small
\caption{Task-specific criteria used by the deterministic construction verifier
and the GRPO accuracy reward. The E-c construction gate is intentionally more
permissive. Final reporting additionally follows the official SemEval metrics
listed in \S\ref{sec:setup}.}
\label{tab:verify-thresholds}
\begin{tabular}{@{}llp{4.0cm}p{4.0cm}@{}}
\toprule
\textbf{Task} & \textbf{Scale} & \textbf{Construction verifier (T)} &
\textbf{GRPO strict criterion} \\
\midrule
EI-reg & $[0,\,1]$  & Absolute error $\leq 0.05$ & Same \\
V-reg  & $[-1,\,1]$ & Absolute error $\leq 0.05$ & Same \\
E-c    & 11 labels  & Label-set $F_1 \geq 0.7$ & Exact label-set match \\
V-oc   & 7 classes  & Exact class match & Same \\
\bottomrule
\end{tabular}
\end{table}

\subsection{Domain-Aware Refinement and Directional Hints}\label{sec:hint-refinement}

When the initial CoT fails verification, the pipeline enters an
iterative refinement loop. The original framework either randomly
selects from four generic search strategies or falls back to
\emph{label leakage}, where the ground-truth answer is injected into the
prompt. Label leakage guarantees convergence at the cost of
superficially correct reasoning that contaminates the training data.

\paragraph{Domain-aware search strategies.}\label{sec:rw-grounding}
The four strategies---\emph{Backtracking}, \emph{Exploring New Paths},
\emph{Verification}, and \emph{Correction}---are each augmented with sentiment- and
emotion-specific guidance grounded in a finding from affective science, so the
refinement mirrors how affect is expressed and resolved in text rather than
following a generic ``rethink''. \emph{Backtracking} fires when a deeper cue
overturns the surface reading: emotion and sentiment differ \cite{munezero2014},
valence shifters (negation, intensifiers, diminishers) can flip or rescale a
valenced word \cite{kennedy2006}, and irony reverses the literal affect
\cite{reyes2012}. \emph{Exploring New Paths} switches from lexical to contextual
cues---idioms and metaphors are non-compositional \cite{prochnow2024}, emotions
co-occur in regular (Plutchik) patterns \cite{jabreel2019deep}, and Russell's
circumplex \cite{russell1980circumplex} spreads affect over arousal $\times$ valence.
\emph{Verification} checks the draft label directly against evidence in the tweet,
including emotion words, shifters, emoji, and punctuation. \emph{Correction} fixes
fine-grained, often ordinal errors---degree adverbs give an intensity order
(\emph{slightly} $<$ \emph{very} $<$ \emph{extremely}) \cite{kennedy2006}, the
anger--disgust co-occurrence \cite{mill2018role} licenses targeted E-c fixes, and the
contextual-versus-prior polarity distinction \cite{wilson2009} guides the V-oc
thresholds. The strategies are applied \emph{one at a time across successive
refinement rounds}: each round randomly selects one, appends its reasoning, and
re-verifies (the inner loop of Stage~3, \S\ref{sec:pipeline}).

\paragraph{Hint mode.}
Instead of revealing the label, the refinement prompt receives only a
\emph{directional hint} derived from the verifier. We design
task-specific hint logic rather than a uniform error-based scheme:
\begin{itemize}[nosep]
  \item \textbf{Regression tasks} (EI-reg, V-reg): the hint indicates
        whether the prediction is \emph{too high}, \emph{too low},
        \emph{slightly high}, \emph{slightly low}, or \emph{close},
        calibrated by error magnitude, with the band cut-offs given in the
        released code; on the signed valence scale these are rendered as
        \emph{too positive}/\emph{too negative}.
  \item \textbf{Multi-label classification} (E-c): the hint conveys
        \emph{over-identified} (too many emotions), \emph{incomplete}
        (missed emotions, with co-occurrence reminders such as
        anger--disgust), or \emph{partially correct}.
  \item \textbf{Ordinal classification} (V-oc): instead of a generic
        ``incorrect'' signal, the hint specifies sentiment polarity
        direction with three granularity levels: \emph{slightly too
        positive/negative} (one-level deviation), \emph{too
        positive/negative} (two- to three-level), and \emph{far too
        positive/negative} (four or more levels).
\end{itemize}
The model is steered toward the correct answer without ever seeing the
label, so the resulting reasoning chain stays authentic.

\paragraph{Failure handling.}
After exhausting the search budget, three mutually exclusive strategies
are available: (1)~\emph{hint refinement}, which provides a directional
hint and re-attempts verification without forcing success (recommended);
(2)~\emph{allow failure}, which marks the sample as unconverged and
excludes it from the SFT set (it is still usable for RL,
\S\ref{sec:quality-tiers}); or (3)~\emph{label leakage} (legacy), which
injects the ground-truth label; such samples are flagged and assigned the
lowest quality tier.

\paragraph{Enhanced search budget.}
We raise the maximum number of fresh restarts from 1 to 3 and the
refinement depth per attempt from 2 to 3, giving the search more room
to converge without resorting to label leakage.

\subsection{Quality Tiering and Data Routing}\label{sec:quality-tiers}

Each generated sample is assigned a quality tier based on its
verification trajectory (Table~\ref{tab:quality-tiers}). The
corresponding quality weight can be used for weighted loss during
supervised fine-tuning or for data selection.

\begin{table}[t]
\centering\small
\caption{Quality tier definitions. \emph{Iterations} denotes the number
of verification rounds before the first success.}
\label{tab:quality-tiers}
\begin{tabular}{@{}lclp{4.4cm}@{}}
\toprule
\textbf{Tier} & \textbf{Wt.} & \textbf{Iter.} & \textbf{Description} \\
\midrule
Gold   & 1.0 & 1     & First-attempt success \\
Silver & 0.8 & 2--3  & Success after limited refinement \\
Bronze & 0.5 & $\geq$4 & Success after extensive refinement \\
Low    & 0.0 & ---   & Success via label leakage \\
Unconv.\ & 0.0 & --- & No correct answer reached \\
\bottomrule
\end{tabular}
\end{table}

Here the quality tier mainly serves as a converged/unconverged filter. Inspired
by the two-stage design of HuatuoGPT-o1~\cite{huatuogpt-o1}, we route verified
reasoning traces to SFT and answer-labelled cases to GRPO. We split each subtask's training
pool 50/50 by a fixed seed (42) into \emph{half-A} and \emph{half-B}. Half-A
is sent through the CoT pipeline above; its \emph{converged} samples
(gold/silver/bronze, all carrying a non-empty reasoning chain and an
answer) form the \textbf{SFT} set. The \textbf{RL} set is the union of
(i)~all of half-B, which is never sent through CoT generation and so
carries only (question, answer) pairs, and (ii)~the \emph{unconverged}
hard samples from half-A. GRPO can still extract a reward signal from a
(question, answer) pair without a CoT trace. Under this routing every
training instance is used exactly once (\S\ref{sec:setup}). Weighted loss using the
gold/silver/bronze sub-weights is left as future work; in the present
setup all retained samples enter training with uniform weight.

\section{Two-Stage Training}\label{sec:training}

We train Qwen3-8B in two stages and defer all hyperparameters to
\S\ref{sec:setup}.

\paragraph{Stage 1: Supervised fine-tuning.}
We fine-tune Qwen3-8B~\cite{qwen3} with full-parameter SFT on the converged,
reasoning-augmented SFT set, teaching the policy to emit a structured
\texttt{<think>} trace followed by an \texttt{<answer>}. This stage fixes the
reasoning format and yields a usable initial policy, whose checkpoint
initialises the RL stage.

\paragraph{Stage 2: Reinforcement learning with GRPO.}
Starting from the SFT policy, we apply GRPO~\cite{shao2024deepseekmath} on the RL
set with a binary, tolerance-matched reward (\S\ref{sec:setup}): a sampled
completion scores $1$ when its parsed answer meets the task tolerance $\tau$ of
\S\ref{sec:task-formulation} and $0$ otherwise. GRPO estimates advantages from
groups of sampled completions, so it needs no separate value network. It learns
directly from the strict criterion of Table~\ref{tab:verify-thresholds}, which
also gates construction on the regression and ordinal subtasks, closing the loop
between synthesis and optimisation. The unconverged hard cases (a gold
answer but no CoT trace) supply additional reward signal at this stage rather
than being discarded.

\section{Experiments and Results}\label{sec:experiments}

\subsection{Experimental Setup}\label{sec:setup}

We run the full pipeline with all four NTDH improvements enabled.

\paragraph{Data.}
We build on
SemEval-2018 Task~1~\cite{mohammad2018semeval}, restricted to the four English
subtasks (E-c, EI-reg, V-reg, V-oc). Each subtask's training pool is halved by a
fixed seed (\S\ref{sec:quality-tiers}): half-A is sent through CoT synthesis and
its converged samples become the SFT set, while half-B together with the
unconverged half-A hard cases form the RL set. Every training instance is
therefore used exactly once (100\% utilisation; 16{,}302 instances total:
5{,}388 SFT and 10{,}914 RL), and Table~\ref{tab:datastats} gives the
per-subtask split sizes.
\begin{table}[t]
\centering\small
\caption{Dataset statistics per subtask. Half-A is sent through CoT
generation; its converged samples form the SFT set, and its hard
(2{,}746 unconverged and 16 empty- or invalid-chain) samples join half-B in the
RL set. All reported
results use the official test split.}
\label{tab:datastats}
\setlength{\tabcolsep}{4pt}
\begin{tabular}{@{}lrrrrr|r@{}}
\toprule
 & \textbf{half-A} & \textbf{SFT} & \textbf{$\to$RL} & \textbf{half-B} & \textbf{RL} & \textbf{official} \\
\textbf{Subtask} & (CoT) & (conv.) & (hard) & (RL src) & (total) & (test) \\
\midrule
EI-reg & 3{,}551 & 2{,}585 & 966   & 3{,}551 & 4{,}517 & 4{,}068 \\
E-c    & 3{,}419 & 1{,}999 & 1{,}420 & 3{,}419 & 4{,}839 & 3{,}259 \\
V-reg  & 590     & 315     & 275   & 591     & 866     & 937 \\
V-oc   & 590     & 489     & 101   & 591     & 692     & 937 \\
\midrule
Total  & 8{,}150 & 5{,}388 & 2{,}762 & 8{,}152 & 10{,}914 & 9{,}201 \\
\bottomrule
\end{tabular}
\end{table}

\paragraph{Realised quality-tier distribution.}
Table~\ref{tab:tier-dist} reports the quality tiers (\S\ref{sec:quality-tiers})
assigned to the 8{,}150 half-A CoT generations, computed from each sample's
verification trajectory. \emph{No} sample required label leakage (the
\emph{low} tier is empty): directional hints (\S\ref{sec:hint-refinement})
replaced the gold-answer injection that the original framework fell back to.

\begin{table}[t]
\centering\small
\caption{Realised quality-tier distribution of the 8{,}150 half-A CoT
generations, assigned by verification trajectory (\S\ref{sec:quality-tiers}).
The gold/silver/bronze tiers contain 5{,}404 converged trajectories; 16
empty- or invalid-chain cases are routed to RL, leaving 5{,}388 SFT
trajectories. The \emph{low} (label-leakage) tier is empty under the
hint-refinement fallback.}
\label{tab:tier-dist}
\begin{tabular}{@{}lrrrrr@{}}
\toprule
\textbf{Task} & \textbf{Gold} & \textbf{Silver} & \textbf{Bronze} & \textbf{Unconv.} & \textbf{Total} \\
\midrule
EI-reg & 371   & 1{,}477 & 744   & 959   & 3{,}551 \\
E-c    & 914   & 828   & 260   & 1{,}417 & 3{,}419 \\
V-reg  & 45    & 186   & 85    & 274   & 590 \\
V-oc   & 169   & 262   & 63    & 96    & 590 \\
\midrule
All    & 1{,}499 & 2{,}753 & 1{,}152 & 2{,}746 & 8{,}150 \\
\bottomrule
\end{tabular}
\end{table}

\paragraph{Baselines.}
We situate our model against the prior systems collected in the EmoLLM
benchmark~\cite{emollm}, in four groups: (i)~the SemEval-2018 competition
systems SeerNet~\cite{duppada2018seernet} and NTUA-SLP~\cite{baziotis2018ntua};
(ii)~fine-tuned pre-trained
language models (BERT, RoBERTa, SentiBERT); (iii)~zero- and few-shot LLM
baselines (Falcon, Vicuna, LLaMA2, ChatGPT, GPT-4); and (iv)~the
instruction-tuned EmoLLM family (EmoBART/T5/OPT/BLOOM and EmoLLaMA), of which
EmoLLaMA-chat-13B is the strongest on EI-reg and V-oc and serves as our
primary reference. All
baseline numbers are reproduced verbatim from~\cite{emollm} and are scored on
the official test split. Table~\ref{tab:vs-semeval} reports their results and
our official-test results in the same task-specific format.

\paragraph{Metrics.}
We report the SemEval-2018 primary metrics, so that our results are directly
comparable with prior work:
\begin{itemize}[nosep]
  \item \textbf{EI-reg, V-reg} (regression) and \textbf{V-oc} (7-class
        ordinal): Pearson correlation $r$ against the gold values, with
        EI-reg macro-averaged over the four target emotions.
  \item \textbf{E-c} (11-label multi-label): Jaccard similarity (the SemEval
        ``acc''), micro-F1, and macro-F1.
\end{itemize}
The strict, tolerance-based accuracy of Table~\ref{tab:verify-thresholds} is
used for the GRPO reward but is not reported as a headline metric, since no
prior system reports it.
Predictions are parsed from the structured \texttt{<think>}/\texttt{<answer>}
output produced by the policy; there are no parsing failures for either
checkpoint reported on the official test.

\paragraph{Models and training.}
We fine-tune \textbf{Qwen3-8B}~\cite{qwen3} with full-parameter SFT,
then run GRPO~\cite{shao2024deepseekmath}
from the SFT checkpoint with the open-r1 / TRL stack~\cite{openr1,trl};
Table~\ref{tab:hyperparams} lists the hyperparameters. We report two
checkpoints on the official test: the SFT checkpoint used to initialise GRPO
(\emph{SFT init}, step~280) and the last saved GRPO checkpoint
(\emph{RL-final}). Throughout, ``step'' denotes the trainer's logged global
step. Both were fixed in advance; no checkpoint selection was performed on the
official test split.

\begin{table}[t]
\centering\small
\caption{Training hyperparameters for the two stages, both fine-tuning
Qwen3-8B. ``---'' marks settings that do not apply.}
\label{tab:hyperparams}
\begin{tabular}{@{}lll@{}}
\toprule
 & \textbf{SFT} & \textbf{GRPO} \\
\midrule
Framework             & LLaMA-Factory        & open-r1 / TRL \\
Learning rate         & $5{\times}10^{-6}$   & $1{\times}10^{-6}$ \\
LR schedule           & cosine, $0.1$ warmup & cosine, $0.1$ min-LR \\
Epochs                & 3                    & 2 \\
Per-device batch      & 1 ($\times 2$ GPUs)  & 2 \\
Gradient accumulation & 8                    & 4 \\
Max length            & 8192                 & 4096 / 8192 (prompt / completion) \\
Precision / runtime   & bf16, ZeRO-3         & vLLM colocated generation \\
Validation split      & $10\%$ held out      & --- \\
Group size $G$        & ---                  & 8 \\
KL coefficient $\beta$ & ---                 & $0.01$ \\
Sampling temperature  & ---                  & $1.0$ \\
\bottomrule
\end{tabular}
\end{table}

\paragraph{Reward function.}
The GRPO reward combines a binary accuracy term with a small format bonus,
$1.0\cdot r_{\text{acc}} + 0.1\cdot r_{\text{fmt}}$. The accuracy
reward $r_{\text{acc}}$ is \emph{binary}: it is $1$ when the parsed prediction
satisfies the task's \emph{GRPO strict criterion} in
Table~\ref{tab:verify-thresholds}---note that this requires an exact label-set
match on E-c, unlike the permissive construction gate---and $0$ otherwise; a
missing \texttt{<answer>} block scores $0$. The format reward $r_{\text{fmt}}$ is $1$ iff the output
matches the \texttt{<think>\ldots</think>\textbackslash n\textbackslash n<answer>\ldots</answer>}
template. The consequences of this binary objective are examined through the
official-test error analysis in \S\ref{sec:discussion}.

\subsection{Main Results and Comparison with Prior Work}\label{sec:vs-sota}

\begin{table}[!t]
\centering\small
\caption{Comparison with prior work on SemEval-2018 Task~1 (English).
EI-reg / V-reg / V-oc use Pearson correlation $r$; the EI-reg ``ave''
column is the macro-average over the four emotions. E-c is reported with
multi-label accuracy (``acc'', i.e.\ Jaccard), micro-F1 and macro-F1.
All systems, including the two NTDH checkpoints, are evaluated on the
official test split. The best score in each column is in \textbf{bold}, and the
second-best score is \underline{underlined}; ties receive the same marking.
EI-oc is excluded from our pipeline by design, so its columns are omitted.}
\label{tab:vs-semeval}
\setlength{\tabcolsep}{3pt}
\resizebox{\textwidth}{!}{%
\begin{tabular}{@{}lccccc|cc|ccc@{}}
\toprule
 & \multicolumn{5}{c|}{\textbf{EI-reg} (Pearson)} & \textbf{V-reg} & \textbf{V-oc} & \multicolumn{3}{c}{\textbf{E-c}} \\
\cmidrule(lr){2-6} \cmidrule(lr){9-11}
\textbf{System} & ave & anger & fear & joy & sadness & (Pearson) & (Pearson) & acc & mi-F1 & ma-F1 \\
\midrule
\multicolumn{11}{@{}l}{\textit{SemEval-2018 competition systems}} \\
SemEval baseline      & 0.520 & ---   & ---   & ---   & ---   & 0.585 & 0.509 & ---   & ---   & ---   \\
NTUA-SLP              & 0.757 & ---   & ---   & ---   & ---   & 0.839 & 0.765 & \underline{0.579} & ---   & ---   \\
SeerNet               & 0.799 & \underline{0.827} & 0.779 & 0.792 & 0.798 & 0.873 & \underline{0.856} & \textbf{0.609} & \textbf{0.724} & \textbf{0.592} \\
\midrule
\multicolumn{11}{@{}l}{\textit{Pre-trained language models}} \\
BERT-base             & 0.785 & 0.800 & 0.781 & 0.783 & 0.742 & 0.840 & 0.805 & 0.567 & 0.718 & \underline{0.568} \\
RoBERTa-base          & 0.717 & 0.670 & 0.736 & 0.769 & 0.694 & 0.845 & 0.772 & 0.563 & \underline{0.721} & 0.536 \\
SentiBERT             & 0.722 & 0.724 & 0.740 & 0.731 & 0.691 & 0.835 & 0.763 & 0.535 & 0.700 & 0.522 \\
\midrule
\multicolumn{11}{@{}l}{\textit{Zero- / few-shot LLM baselines}} \\
Falcon                & 0.114 & 0.147 & 0.082 & 0.095 & 0.131 & 0.135 & 0.189 & 0.190 & 0.318 & 0.253 \\
Vicuna                & 0.281 & 0.307 & 0.257 & 0.260 & 0.299 & 0.298 & 0.579 & 0.220 & 0.359 & 0.253 \\
LLaMA2-7B-chat        & 0.194 & 0.176 & 0.257 & 0.097 & 0.247 & 0.094 & 0.497 & 0.257 & 0.414 & 0.286 \\
LLaMA2-13B-chat       & 0.488 & 0.524 & 0.506 & 0.398 & 0.526 & 0.312 & 0.568 & 0.274 & 0.424 & 0.302 \\
ChatGPT               & 0.599 & 0.637 & 0.573 & 0.569 & 0.618 & 0.637 & 0.748 & 0.382 & 0.546 & 0.429 \\
ChatGPT-FS            & 0.550 & 0.572 & 0.482 & 0.587 & 0.560 & 0.739 & 0.791 & 0.413 & 0.563 & 0.466 \\
GPT-4                 & 0.656 & 0.699 & 0.575 & 0.686 & 0.667 & 0.811 & 0.788 & 0.444 & 0.572 & 0.497 \\
GPT-4-FS              & 0.679 & 0.704 & 0.654 & 0.679 & 0.678 & 0.825 & 0.793 & 0.460 & 0.582 & 0.515 \\
\midrule
\multicolumn{11}{@{}l}{\textit{EmoLLM family (instruction tuning on AAID)}} \\
EmoBART               & 0.795 & 0.798 & 0.803 & 0.795 & 0.782 & 0.851 & 0.835 & 0.528 & 0.686 & 0.548 \\
EmoT5                 & 0.783 & 0.785 & 0.797 & 0.798 & 0.751 & 0.852 & 0.836 & 0.559 & 0.712 & \underline{0.568} \\
EmoOPT                & 0.825 & \underline{0.827} & 0.830 & 0.837 & 0.805 & \textbf{0.887} & 0.843 & 0.532 & 0.680 & 0.550 \\
EmoBLOOM              & 0.791 & 0.802 & 0.797 & 0.790 & 0.776 & 0.857 & 0.822 & 0.528 & 0.683 & 0.552 \\
EmoLLaMA-7B           & 0.822 & 0.819 & 0.821 & 0.837 & 0.809 & 0.879 & 0.843 & 0.545 & 0.695 & 0.563 \\
EmoLLaMA-chat-7B      & 0.824 & 0.825 & 0.830 & 0.832 & 0.810 & 0.876 & 0.827 & 0.534 & 0.693 & 0.540 \\
EmoLLaMA-chat-13B     & \underline{0.831} & \underline{0.827} & \underline{0.835} & 0.843 & \underline{0.817} & \underline{0.886} & \textbf{0.860} & 0.537 & 0.696 & 0.545 \\
\midrule
\rowcolor{gray!15}\multicolumn{11}{@{}l}{\textit{Ours} (NTDH CoT $\to$ SFT $\to$ GRPO on Qwen3-8B)} \\
\rowcolor{gray!15} SFT init (ckpt-280) & 0.800 & 0.753 & 0.799 & \underline{0.845} & 0.802 & 0.785 & 0.785 & 0.557 & 0.667 & 0.555 \\
\rowcolor{gray!15} RL-final            & \textbf{0.862} & \textbf{0.829} & \textbf{0.858} & \textbf{0.879} & \textbf{0.882} & 0.840 & 0.831 & \underline{0.579} & 0.677 & 0.543 \\
\bottomrule
\end{tabular}%
}
\end{table}

\paragraph{Official-test comparison.}
The official NTDH scores and prior-system results are compared directly in
Table~\ref{tab:vs-semeval}. NTDH achieves its strongest result on EI-reg.
RL-final obtains a macro-averaged Pearson correlation of $0.862$, outperforming
SeerNet by $0.063$ and EmoLLaMA-chat-13B by $0.031$. It also achieves the
highest correlation on all four target emotions. Performance is more mixed on
the other subtasks. RL-final reaches
$0.840$ on V-reg and $0.831$ on V-oc, trailing the strongest baselines by
$0.047$ and $0.029$, respectively, while its E-c Jaccard of $0.579$ is the
second-best score in the table.

\paragraph{Comparison within the EmoLLM family.}
NTDH remains competitive with the EmoLLM family while using approximately
$14\times$ fewer total training records than EmoLLM's AAID corpus. RL-final exceeds all seven
EmoLLM variants on EI-reg, while its V-reg and V-oc results remain below the
strongest members of that family. On E-c it achieves a Jaccard score of
$0.579$, exceeding the strongest EmoLLM result, EmoT5 at $0.559$, by $0.020$,
while remaining $0.030$ below SeerNet.
This pattern indicates that the compact NTDH corpus is effective for
emotion-intensity regression and structured multi-label prediction, although
macro-F1 shows that rare-label recall remains a limitation.

\paragraph{Two-stage training outcome.}
Relative to the SFT initialisation checkpoint on the same official split, RL-final
improves EI-reg Pearson correlation from $0.800$ to $0.862$, V-reg from $0.785$
to $0.840$, and V-oc from $0.785$ to $0.831$. On E-c, Jaccard increases from
$0.557$ to $0.579$ and micro-F1 from $0.667$ to $0.677$, whereas macro-F1
decreases from $0.555$ to $0.543$. The EI-reg improvement occurs across all
four target emotions: $+0.076$ for anger, $+0.059$ for fear, $+0.034$ for joy,
and $+0.080$ for sadness.

\subsection{Ablation of NTDH Components}\label{sec:ablation}

We isolate each NTDH component and measure its effect on the synthesis output,
scored against the true tolerance $\tau$ of \S\ref{sec:task-formulation}. These
figures are data-side: they characterise the quality of the synthesised data
independently of downstream model evaluation. For N, T and H, the effect is an \emph{offline
counterfactual} read off the full synthesis run; for D, which is on by default, we run a matched $320$-instance re-synthesis with generic rethink prompts.
Table~\ref{tab:ablation} summarises the results.

\begin{center}
\begin{minipage}{\linewidth}
\centering\small
\makeatletter\def\@captype{table}\makeatother
\caption{Ablation of the four NTDH components, each removed in isolation. Metrics
are data-side, measured on the CoT-synthesis output against the true tolerance.
For N, $18.4\%$ is the proportion of initial model conclusions within gold
tolerance over all 8{,}150 half-A samples; $100\%$ denotes correctness of the
gold-derived answer field for retained SFT traces. T and H are computed on the
full run and D on a matched $320$-instance sample.}
\label{tab:ablation}
\begin{tabular}{@{}llcc@{}}
\toprule
\textbf{Component removed} & \textbf{Data-quality metric} & \textbf{NTDH} & \textbf{w/o} \\
\midrule
Naturalisation (N) & answer-source gold consistency       & $100\%$       & $18.4\%$ \\
Tolerance gate (T) & out-of-tol.\ accepts (EI-reg / V-reg) & $0\%$         & $62.7\% / 47.5\%$ \\
Domain strat.\ (D) & convergence yield (overall / E-c)     & $66\% / 56\%$ & $65\% / 46\%$ \\
Hints (H)          & label leakage in the fallback         & $0\%$         & gold answer injected \\
\bottomrule
\end{tabular}
\end{minipage}
\end{center}

\paragraph{Naturalisation (N).}
Without naturalisation, the candidate \texttt{<answer>} is the model's own
initial conclusion, which lands within gold tolerance for only $18.4\%$ of the
8{,}150 half-A samples, and far less often on the regression subtasks. With
naturalisation, the answer source is instead the naturalised gold label, so the
\texttt{<answer>} field of every retained SFT sample is correct by construction.
Accordingly, $100\%$ here denotes answer-target correctness. The synthesis stage
retains 5{,}388 samples with verified, non-empty traces for SFT and routes the
other 2{,}762 half-A samples to RL.

\paragraph{Tolerance gate (T).}
Replacing the deterministic gate with the original LLM judge silently admits
out-of-tolerance regression answers most of the time (Table~\ref{tab:ablation}),
corrupting the supervised targets; the ordinal and multi-label subtasks are less
affected. The deterministic gate removes this leakage entirely.

\paragraph{Domain-aware strategies (D).}
Domain strategies barely move the overall convergence yield, but the effect is
task-dependent: they help the fine-grained, multi-label E-c subtask by about
$10$ points (Table~\ref{tab:ablation}), where Plutchik co-occurrence and the
circumplex apply, while leaving the coarser regression subtasks unchanged. Their
role is to keep the refinement grounded in affective-science theory rather than
to raise yield, consistent with reasoning mattering most at the micro level.
These are single-run estimates, so the small overall gap may reflect run-to-run
variation; the larger E-c difference is promising but requires repeated runs
before it can be treated as robust.

\paragraph{Directional hints (H).}
When a trace exhausts the search budget, the original framework fell back to
injecting the gold answer, so any trace rescued that way carried the target in
its own reasoning. Directional hints replace that fallback with a label-free
signal, and the effect is visible in the realised tier distribution: the
\emph{low} (label-leakage) tier of Table~\ref{tab:tier-dist} is empty, so
\emph{no} retained trace owes its supervision to a leaked answer
(Table~\ref{tab:ablation}). Traces that the hints do not rescue are routed to RL
rather than repaired, so the mechanism trades a lower yield for supervision that
is honestly derived throughout.

\section{Discussion}\label{sec:discussion}

\paragraph{Target quality under sparse verifiable rewards.}
The final two-stage policy scores above the SFT initialisation checkpoint on five of
six official-test metrics while using 16{,}302 total SFT and RL records,
approximately $14\times$ fewer than the $\sim$$234$K records in EmoLLM's AAID
corpus. The SFT stage itself uses only 5{,}388 reasoning trajectories.
This result is consistent with the importance of target quality when reward is
sparse. With a $\{0,1\}$ reward and group-relative advantages, a
prompt group yields no GRPO gradient when all its completions score alike. A
controlled estimate obtained with a proxy generator rather than the fine-tuned
policy ($n=70$ prompts, $k=8$ samples per prompt) places the share of such
zero-variance groups at $77.1\%$ for E-c and $75.7\%$ for V-reg; RL thus
learns only from the minority of groups that straddle the decision boundary, while
wrong SFT targets anchor the policy badly before RL even begins. Naturalisation
replaces the model-derived answer source (only $18.4\%$ of initial conclusions
meet the gold tolerance) with gold-consistent targets for every retained SFT
sample. Together with the tolerance-aware gate, it is therefore a main lever;
adding data mostly adds more zero-variance
groups. The one official-test metric that declines, E-c macro-F1
($0.555\!\rightarrow\!0.543$), motivates the qualitative error analysis below.

\paragraph{Error analysis on the official test.}
The representative cases in Table~\ref{tab:error-cases} expose three recurring
failure modes: reasoning--answer inconsistency on E-c, collapse of
mixed-valence clauses into neutrality on V-oc, and over-correction on explicit
sarcasm markers. The remaining errors therefore arise from rare-label omission,
fine-grained intensity calibration, and conflicting contextual cues rather than
from output formatting.

\paragraph{Limitation of permissive E-c verification.}
The $F_1\geq0.7$ E-c construction threshold improves trace coverage but does
not guarantee that an accepted trace supports every label in the complete
gold answer supplied through Naturalisation. Consequently, some E-c SFT
records may contain a locally plausible reasoning trace paired with a broader
gold label set. Our present verifier checks the candidate conclusion against
the gold set but does not separately test entailment between each reasoning
step and each output label. The results should therefore not be interpreted as
evidence that every generated rationale is fully faithful. A label-wise
reasoning--answer consistency gate is an important next step.

\begin{table}[t]
\centering\small
\caption{Representative errors of the GRPO policy on the official test. Gold and predicted
labels are shown in the benchmark's output space; excerpts are shortened only
for presentation.}
\label{tab:error-cases}
\begin{tabular}{@{}lp{4.0cm}p{3.4cm}p{5.0cm}@{}}
\toprule
\textbf{Task} & \textbf{Tweet excerpt} & \textbf{Gold $\rightarrow$ prediction} & \textbf{Observed failure} \\
\midrule
E-c & ``Indian army is on its way to dispatch all terrorists to hell''
& \{anger, anticipation, optimism, trust\} $\rightarrow$ \{anger, disgust\}
& The trace mentions anticipation and optimism, but the final answer omits
them and substitutes disgust, exposing a reasoning--answer consistency error. \\
V-oc & ``YAY, that's awesome \ldots\ it makes me sad you're leaving''
& $-2 \rightarrow 0$
& Opposing positive and negative clauses are collapsed into neutrality instead
of resolving which cue controls the ordinal judgement. \\
V-reg & ``The fun never stops. [upside-down-face] \#sarcasm''
& $0.392 \rightarrow -0.266$
& The explicit sarcasm marker triggers a polarity reversal stronger than the
benchmark annotation, illustrating over-correction by a salient domain cue. \\
\bottomrule
\end{tabular}
\end{table}

\paragraph{Implications of the cases.}
The E-c example shows that a plausible trace is insufficient unless the final
answer is checked against the trace itself; future verification should therefore
add a reasoning--answer consistency test before applying the gold-based gate.
The mixed-valence V-oc case motivates an explicit cue-weighting step, whereas
the sarcasm case cautions against treating a domain hint as a deterministic
label flip. Together, the cases favour confidence-aware, label-specific rewards
over uniformly stronger refinement.

\paragraph{Two lessons for verifiable-reward pipelines.}
First, the binary reward decouples strict accuracy from the continuous metrics
by construction: once a prediction enters the $\pm0.05$ band, no additional
credit distinguishes a close estimate from an exact one. A shaped reward (soft
F1 for E-c and distance-aware credit for regression and ordinal tasks) is
therefore a natural next step. Second, multi-stage
SFT$\rightarrow$RL routing changes dataset membership after the initial split,
so split disjointness has to be re-verified after every routing decision rather
than once at the start.

\section{Conclusion}\label{sec:conclusion}

We asked whether a small, carefully synthesised reasoning corpus plus RL can
make an 8B model competitive on fine-grained affective analysis without the
hundreds of thousands of supervised instances prior recipes use. NTDH answers
the data-quality half of the question: ground-truth naturalisation and a
tolerance-aware gate provide gold-consistent answer targets for all 5{,}388
retained SFT traces, while routing the other 2{,}762 half-A cases to RL uses all
$16{,}302$ instances. A
component ablation (\S\ref{sec:ablation}) isolates the contribution of each
stage. On the official 9{,}201-instance test
set, RL-final obtains Pearson correlations of $0.862$, $0.840$, and $0.831$ on
EI-reg, V-reg, and V-oc, respectively, together with Jaccard, micro-F1, and
macro-F1 scores of $0.579$, $0.677$, and $0.543$ on E-c. It scores above the
SFT initialisation checkpoint on five of six metrics and achieves the strongest
EI-reg result among the compared systems, demonstrating competitive performance
across four heterogeneous subtasks
(Table~\ref{tab:vs-semeval}). Two refinements follow from our analysis: a
shaped GRPO reward that grades how close a prediction lies to the scoring
tolerance, and a quality-weighted SFT loss that uses the gold/silver/bronze
tiers already produced during synthesis.

Beyond SemEval-2018, NTDH is a general recipe for building verifiable-reward
reasoning data. Its two core ideas (deriving the verification gate and the
reward from one task tolerance, and steering failed traces with label-free
directional hints) apply to any task with a checkable answer, such as graded
relevance, ordinal stance, or multi-label tagging. NTDH therefore applies to verifiable-reward training on
information-retrieval and text-mining tasks, not only on affect.

\section*{Code and Data Availability}
The NTDH data pipeline, the (frozen) SFT/GRPO training configurations,
the evaluation harness, the released prediction and metric
files, and the official-test inference script are organised in a single
self-contained repository, to be released publicly upon publication. The two
training files (SFT and RL) rebuild
bit-identically from the raw SemEval-2018 inputs at the fixed seed
reported above.

\section*{Ethical Considerations}
SemEval-2018 Task~1 is a public benchmark of English tweets, used here
under its original terms; the excerpts in Table~\ref{tab:error-cases} are
included solely for aggregate error analysis and are minimally reproduced.
Affective analysis of social-media text carries dual-use risk: the same
models that can support, e.g., mental-health screening can also enable
intrusive surveillance of individuals' emotional states. We therefore
frame this work as a study of \emph{reasoning and data quality} on a
fixed academic benchmark, and caution that any deployment on real users
requires consent, privacy safeguards, and domain-appropriate validation
well beyond the benchmark evidence presented here.

\bibliographystyle{model1-num-names}
\bibliography{refs}

\end{document}